\def\year{2022}\relax
\documentclass[letterpaper]{article} % DO NOT CHANGE THIS
\usepackage{aaai22}  % DO NOT CHANGE THIS
\usepackage{times}  % DO NOT CHANGE THIS
\usepackage{helvet}  % DO NOT CHANGE THIS
\usepackage{courier}  % DO NOT CHANGE THIS
\usepackage[hyphens]{url}  % DO NOT CHANGE THIS
\usepackage{graphicx} % DO NOT CHANGE THIS
\usepackage[T1]{fontenc}
\usepackage{natbib}  % DO NOT CHANGE THIS AND DO NOT ADD ANY OPTIONS TO IT
\usepackage{caption} % DO NOT CHANGE THIS AND DO NOT ADD ANY OPTIONS TO IT
\DeclareCaptionStyle{ruled}{labelfont=normalfont,labelsep=colon,strut=off} % DO NOT CHANGE THIS
\usepackage{algorithm}
\usepackage{algorithmic}
\usepackage{pdflscape}
\usepackage{subcaption}
\usepackage{rotating}

\usepackage{newfloat}
\usepackage{listings}
\floatstyle{ruled}
\newfloat{listing}{tb}{lst}{}
\floatname{listing}{Listing}

\title{Discovering collective narratives shifts in online discussions}
\author{
    Wanying Zhao\textsuperscript{\rm 1},
    Fiona Guo\textsuperscript{\rm 2,3},
    Kristina Lerman\textsuperscript{\rm 2,3},
    Yong-Yeol Ahn\textsuperscript{\rm 1,4}
}
\affiliations{
    \textsuperscript{\rm 1}Center for Complex Networks and Systems Research, Luddy School of Informatics, Computing, and Engineering, Indiana University\\
    \textsuperscript{\rm 2}Department of Computer Science, University of Southern California\\
    \textsuperscript{\rm 3}Information Sciences Institute, University of Southern California\\
    \textsuperscript{\rm 4}Network Science Institute, Indiana University, Bloomington, IN, USA\\
    \{zhaowany, yyahn\}@iu.edu\\
    \{siyiguo, lerman\}@isi.edu
}

\usepackage{bibentry}
\begin{document}

\maketitle

\begin{abstract}
Narrative is a foundation of human cognition and decision making. Because narratives play a crucial role in societal discourses and spread of misinformation and because of the pervasive use of social media, the narrative dynamics on social media can have profound societal impact. Yet, systematic and computational understanding of online narratives faces critical challenge of the scale and dynamics; how can we reliably and automatically extract narratives from massive amount of texts? How do narratives emerge, spread, and die?
Here, we propose a systematic narrative discovery framework that fill this gap by combining change point detection, semantic role labeling (SRL), and automatic aggregation of narrative fragments into narrative networks. We evaluate our model with synthetic and empirical data---two Twitter corpora about COVID-19 and 2017 French Election. Results demonstrate that our approach can recover major narrative shifts that correspond to the major events. 

%Other than that, we explore the potentials of statistical measurement and network-based method under the context of narrative networks. Both methods help to identify core narratives in massive narrative fragments. In sum, our method works successfully in discovering collective narrative shift in online discussion, and we argue that it could be applied in assorted context, including but not limited to political campaign and conspiracy theories. 
\end{abstract}

\noindent 
Human beings are ``storytelling animals''~\cite{gottschall2012storytelling}. Narratives helps us make sense of our experiences of reality~\cite{richardson1997unlikely, bruner1987life, roberts2001history}, which enriches our emotions (e.g.~\citealt{kopfman1998affective}), influences our beliefs (e.g.~\citealt{appel2007persuasive, de2012identification}), and affects our decisions~\cite{winterbottom2008does}. 

The online social media opened up a global public space where narratives emerge, mutate, compete, spread, and die. 
The online narratives are collectively created by numerous participants~\cite{meraz2013networked, papacharissi2016affective} and can quickly spread globally. 
The potency of online narratives in society (cf.~``Make America Great Again'' or ``Black Lives Matter'') calls for a thorough investigation into how narratives evolve on social media. 

However, it is challenging to study online narrative evolution. 
First, it is a messy, large-scale process involving millions of people.  
Second, narrative is often \emph{fragmented} inside short messages. For instance, Twitter's 140-character limit (expanded to 280 in 2017) leaves little space for users to flush out a narrative~\cite{hermida2014twitter, sadler2018narrative}. 
This makes studying online narratives, particularly from tweets, challenging. 
Second, close reading~\cite{moretti2000conjectures} is labor intensive and does not scale well. 
Third, the concept of narrative is not well-defined and often difficult to operationalize. 

Tackling these challenges, we ask: 
\begin{itemize}
\item \textbf{RQ1}: How can we extract narrative fragments hidden in short posts?
\item \textbf{RQ2}: How can we aggregate and synthesize narrative fragments into collective narratives?
\item \textbf{RQ3}: How can we systematically detect the collective narrative shifts from a massive collection of social media conversations?
\end{itemize}

Our computational framework addresses these research questions in the following ways. 
First, we use change point detection~\cite{he2022leveraging} to identify \emph{collective narrative shifts}. 
Second, we operationalize \emph{narrative fragments}~\cite{dourish2018datafication} (or \emph{narrative elements}~\cite{jing2021characterizing}), using Semantic Role Labeling (SRL)~\cite{shi2019simple}.
Third, we aggregate narrative fragments into a \emph{narrative network} that captures more complete narratives. 
Now, let us introduce these three primary components of our method. 

\paragraph{Narrative fragment discovery}
The most common approaches to narrative discovery are keyword-based and topic-modeling methods. Keyword-based focuses on sets of coherent individual tokens, or keywords  (e.g.,~\citealt{lazard2015detecting}), as a narrative; Topic-modeling discovers ``topics''---another proxy of narrative---each of which is a probability distribution over tokens. Latent Dirichlet Allocation~\cite{blei2003latent} and its variations~\cite{8607040,8571225}, including a BERT-based model~\cite{grootendorst2022bertopic}, and matrix factorization methods like NMF~\cite{lee2000algorithms} are commonly used. 
However, despite the great utility of these methods and the fact that keywords and topics are crucial elements of a narrative, they are still not equivalent to narratives. 

One useful way to conceptualize how narratives unfold in social media is to consider each post as a \emph{chronicle}; i.e., from a stream of events with timestamps (posts), which contain actors, motives, and narrative fragments~\cite{dourish2018datafication}, although they may or may not have direct logical or causal relationships between each other (cf.,~\cite{sadler2018narrative, sadler2021fragmented}).

This is intimately related to \emph{semantic role labeling (SRL)}, which identifies triplets of Action (a verb), Agent (who initiates the action) and Patient (the recipient of the action) from a sentence~\cite{fillmore1967case}. 
For example, the sentence ``I love this coffee shop'', contains a triplet of [``love'' (action), ``I'' (agent), ``coffee shop'' (patient)]. 
SRL co-advanced with the advancement of large language models, leading to powerful tools, such as SENNA~\cite{collobert2011natural} and AllenNLP~\cite{gardner2018allennlp}. 
Indeed, SRL was shown to be able to extract relevant narrative fragments from politician's tweets during the early stage of the pandemic~\cite{jing2021characterizing}; Similar methods were applied to extract actors and actions from blog posts and news articles about conspiracies~\cite{tangherlini2020automated, shahsavari2020conspiracy}.
Following these studies, we operationalize a narrative fragment as a SRL triplet---(agent, action, patient).

\paragraph{Change point detection}
To understand the emergence and shifts in collective narrative, we need to study the temporality of collective narratives. 
The most basic aspect is the identification of significant inflection points where shifts in collective narratives happen.

Many research studied the detection of significant change points.  
For example, cumulative summation (CUSUM)~\cite{page1954continuous} is a basic method that detects changes in time by assuming a normality of data distribution. 
However, this idea  (e.g.~\citealt{willsky1976generalized, barber2015generalized}) tends to be limited to univariate time series data and only detects single change point. Alternative methods built on statistical models, such as hidden Markov model~\cite{raghavan2013hidden} and Bayesian inference~\cite{niekum2015online, wilson2010bayesian} tend to provide higher-quality results but at the cost of computational complexity. 
Meta Change Point Detection (MtChD)~\cite{he2022leveraging}
is argued to be a state of the art method that handles sparse and noisy data well. It is computationally efficient and can handle high-dimensional datasets, while maintaining high accuracy. This study thus employs MtChD as the method for change point detection. 

\paragraph{Narrative Network}
After identifying temporal contexts (the time periods defined by the change points) and extracting the narrative fragments that are prevalent in each period, we further synthesize these fragments to create \emph{narrative networks}. The idea was originally proposed by Bearman and Stovel~\cite{bearman2000becoming}, where they argued that network representation can link pieces of a narrative together. Although the idea was initially used to study individuals' life stories, it can also be used to study public opinions. Past research has shown that by constructing a narrative network, one could form a full story-line of the Bridgegate and Pizzagate~\cite{tangherlini2020automated}. Furthermore, another study has shown the potential of applying community detection algorithm~\cite{blondel2008fast} on narrative networks to uncover major COVID-19 topics~\cite{shahsavari2020conspiracy}. Here, we construct narrative networks based on the narrative fragments extracted by SRL, then apply backbone filtering~\cite{serrano2009extracting} to extract and visualize the core of narrative networks. We also compare this approach with simpler statistical methods.  

Now, we will describe our methods and lay out our findings from two Twitter datasets about COVID-19 and the 2017 French Election. We find that our method accurately captures the major events related to COVID-19, and identify collective narrative shifts associated with Le Pen and Macron following significant change points, which align with the election-related events. Additionally, we evaluate the robustness of our method with synthetic data and baseline methods. 

\section{Methods}
%The overall pipeline of our approach is shown in Figure.~\ref{pipeline}. We will explain each part below. 

%\begin{figure}[t]
%\centering
%\includegraphics[width=\columnwidth]{figures/flow_chart.png}
%\caption{Automated narrative detection pipeline.}
%\label{pipeline}
%\end{figure}

\subsection{Data}
We use two Twitter datasets: a COVID-19 corpus~\cite{chen2020covid} and a 2017 French Election corpus~\cite{ferrara2017disinformation}. 
They cover different topics (politics and pandemic) and two languages (primarily in English and in French, respectively). 
Furthermore, the French election dataset features two key entities (i.e.,~Macron vs. Le Pen) while the COVID-19 corpus is not strongly anchored to contrasting entities. 
Apart from empirical data, we generate three versions of synthetic data to address 1) the robustness of the aggregation method and 2) the robustness of change point detection against noise, and 3) against overlap of collective narratives in time. 
 
\subsubsection{COVID-19 corpus}
The COVID-19 corpus contains tweets about the COVID-19 pandemic collected by using keywords~\cite{chen2020covid}, capturing 600k tweets in English between January 21 to March 31, 2020. 

\subsubsection{2017 French Election corpus}
This corpus includes keyword-identified tweets related to the 2017 French presidential Election~\cite{ferrara2017disinformation}, and covers the period from April 26 to May 29, 2017. We focus on original tweets written
in top three languages in the corpus---French, English, and Spanish, totaling in 2,438K tweets. We translate non-English tweets to English using the MarianMT model\footnote{\url{https://huggingface.co/docs/transformers/model_doc/marian}}.  

We further create two separate, keyword-based French Election corpus---a 1) Macron corpus and a 2) Le Pen corpus. To identify the keywords referring to the candidates, we first trained a word2vec~\cite{church2017word2vec} model using the whole corpus. Then, we select top 20 keywords with the smallest cosine distance to each candidate's name. After that, we iterate the same process for these top 20 keywords until the search was exhaustive. With the potential list of keywords, we manually examine and remove terms that do not refer to one of the candidates. 

% maybe we could skip the detail?
% The final terms that we keep are, for ``Macron'', ``macron'', ``e.macron'', ``emmanuelmacron'', ``m.macron'', ``emacron'', ``emmanuel'', ``macro'', and ``macron2017'', and for ``LePen'', ``marine-le-pen'', ``lepen'', ``mlepen'', ``m.lepen'', ``marine2017'', ``pen2017'', ``lepen2017'', ``le pen'', ``pen'', ``marine lepen'', ``marine le pen'', and ``mlp''. We ignore cases for all terms. 

\subsection{Change Point Detection}
We use MtChD~\cite{he2022leveraging} \footnote{\url{https://github.com/yuziheusc/confusion_multi_change}}, a self-supervised change point detection method based on the concept of confusion-based training~\cite{van2017learning}, which considers changes in classification accuracy as indicators of change points. This method works well on high-dimensional textual data, and it is robust to sparse signals and noise. This is very suitable for the social media data in our study. Specifically, given a randomly selected time point in the data as a trial change point, we train a classifier to predict each data point as before or after the trial change point. If we perform this prediction task at many different time points in the data, we should expect significant changes in the accuracy of this prediction task at true change points. Therefore, the method detect changes as differences in accuracy compared to a null model (predicting the majority class). To construct the classifier, we use TF-IDF with 5,000 most frequent words as the text embedding and compare two types of classification methods: a random forest and a neural network. The change points detected using both methods are very similar. Here we present the results using a random forest. In addition, MtChD also allows us to detect multiple change points by recursively partitioning data on discovered change points. The recursive segmentation has at most three levels and a minimum time length of four days.

\subsection{Semantic Role Labeling (SRL)}
Because SRL operates on sentences, we first split tweets into sentences using the NLTK tokenizer~\cite{loper2002nltk}. Then, we adopt a re-implementation\footnote{\url{https://github.com/Riccorl/transformer-srl}} of the BERT-based SRL model~\cite{shi2019simple} to extract narrative fragments. This model can further predict semantic roles of verbs captured in a sentence (i.e., the function and semantic meaning of a verb in the sentence as in Propbank standard\footnote{\url{https://propbank.github.io/}}\cite{kingsbury2003propbank}), which is missing in similar model provided by AllenNLP\footnote{\url{https://docs.allennlp.org/models/main/models/structured_prediction/predictors/srl/}}. For each sentence, the model generates multiple triplets, each of which represents a narrative fragment in the form of a triplet (Argument 0 \textbf{A0}, Verb \textbf{V}, Argument 1 \textbf{A1}). A0 represents the agent of an action (Verb); A1 refers to the patient or theme of the Verb. This triplet altogether conveys ``who did what'' and ``what did what,'' which reveals the relationship between entities, such as (``you,'' ``vote,'' ``Macron'')---which captures the narrative fragments about `your' voting for Macron (e.g., ``You should vote for Macron''). 

\subsection{Narrative fragment aggregation}

We adopt a three-step process to aggregate narrative triplets (fragments). First, we map semantic roles from Propbank to VerbAtlas\footnote{\url{https://verbatlas.org/}}, which helps compressing 5,649 verbs into 466 verb frames. For example, (``You,'' ``back.01,'' ``Macron'') and (``You,'' ``endorse.01,'' ``Macron'') can be aggregated into (``You,'' ``FOLLOW\_SUPPORT\_SPONSOR\_FUND,'' ``Macron''). With this mapping, we convert 422,019 unique narrative triplets (A0, Verb, A1) into 418,554 triplets (A0, (verb) frame, A1) in COVID data; and 2,016,058 unique ones into 1,253,182 in the French Election data. 

Second, we consider a narrative triplet as a sentence (i.e. ``A0 frame A1'') and train its embedding with a pre-trained SentBert\footnote{\url{https://www.sbert.net/index.html}} model (``all-mpnet-base-v2'') to cluster them. We validate our approach---considering each triplet as a sentence---with the SemEval STS data\footnote{\url{https://github.com/facebookresearch/SentEval}}. We first select sentence pairs with reference similarity (ranges from 0 to 5, determined by human annotators from SemEval) at least four. We consider sentence pairs with high values indicating high semantic similarity. Then, we generate two sets of embeddings by the SentBert model, one on raw SemEval STS sentences, and the other on narrative triplets generated from raw SemEval STS sentences. The test result indicates that the narrative triplets (311/605) even outperform the raw sentences (230/605) in predicting sentence similarity.

We then apply BIRCH~\cite{zhang1996birch} to cluster narrative triplet embeddings. The most frequent triplet in each cluster is considered the representative triplet. 

Finally, we cluster A0 and A1 in the narrative triplets similarly to the second step. In sum, we convert the COVID data into 261,823 unique narrative triplets, and the French Election data into 738,737 unique triplets.   

\subsubsection{Narrative Significance}
After aggregating narrative triplets, we calculate the relevance of each narrative triplet within the time frames discovered by MtChD to identify the most important narrative fragments in each time frame. 
Based on the previous studies~\cite{monroe2008fightin,jing2021characterizing}, we adopt the log-odd ratios with informative Dirichlet priors~\cite{monroe2008fightin} to estimate the relevance of each triplet:
\begin{equation} 
    s_w = \log \frac{f^T_{t}+f^{B}_{t}}{n^T+n^B-f^T_t+f^B_t}-\log \frac{f^R_{t}+f^{B}_{t}}{n^R+n^B-f^R_t+f^B_t}, \nonumber
\end{equation}
where $f^C_{t}$ stands for frequency of a narrative triplet $t$ in corpus $C$; $T$, $R$, and $B$ refers to the target, reference, and background corpus respectively. 
$n^C$ is the size of the corpus $C$ in terms of the number of triplets.
In the French Election data, for example, given the discrepancy between two candidates is more valuable, $T$ would be Macron corpus within a time frame, $R$ would be Le Pen corpus within the same interval, and $B$ would be the whole French Election corpus within the same interval. In the case of the COVID data,  $R$ could be the data of previous time frame and $B$ is the whole COVID corpus. 

\subsubsection{Narrative Network Construction}
We define a narrative network as a weighted, directed network where nodes are arguments (A0 and A1), edges represent verb frames, and the weight of edges is the aforementioned log-odd ratio value. To further illustrate the advantages of applying MtChD, we first construct a global network with overall narrative triplets, and compare it to local networks of significant time periods for both Twitter data. This way, we can better map out how narratives shift across time in the local networks, which is unavailable in the global network.

On top of networks, we apply the multiscale backbone filtering~\cite{serrano2009extracting} 
% schmidt2017anatomy, jiang2020political} 
to extract the core narrative network. 
To focus on the most important narratives, we set $\alpha$ to $10^{-7}$ for the global network and 0.01 for local networks for the COVID data; for the French Election data, we set $\alpha$ to 0.001 for Le Pen related local networks, and 0.01 for Macron related local networks.

\subsection{Evaluation}
We quantitatively evaluate our framework on the two parts: MtChD and SRL. First, we use synthetic data to evaluate MtChD, and then use both synthetic data and the two Twitter corpora to evaluate SRL and clustering performance. 

\subsubsection{Evaluation with Synthetic data}
The key idea in this evaluation is to construct a dataset with planted narratives that change over time by using paraphrasing capacity of a generative language model. 
With these planted narratives and change points that we can manipulate, we test the performance of our method in recovering the change points and core narrative from the synthetic data. 

\subsubsection{Synthetic data generation}

Our synthetic data construction starts with an empirical event data. 
We consider the 22 major events related to COVID-19 provided by The American Journal of Managed Care (AJMC)\footnote{\url{https://www.ajmc.com/view/a-timeline-of-covid19-developments-in-2020}} as the \emph{reference events}, and paraphrase their summary descriptions into ten additional versions using ChatGPT, resulting in a total of 197 reference narratives. 
Additionally, we sample \emph{noise events} from Wikipedia. These noise events are major events happened between 2021 and 2022\footnote{\url{https://en.wikipedia.org/wiki/2021}; \url{https://en.wikipedia.org/wiki/2022}}, such as ``The 2022 Winter Olympics begin in Beijing, China.'' 
We randomly sample 100 events that are not directly related to COVID-19. 
Each noise event is represented in the synthetic data as a single narrative without additional paraphrased sentences. With narrative sentences, SRL extracts 104 narrative fragments in total. As we will explain below, using these basic narrative elements, we generate different synthetic dataset for each task. 

\subsubsection{Evaluation of change point detection} \label{eval_MtChD}
We use synthetic data to evaluate the robustness of MtChD against noise and temporal overlap by comparing its performance with a widely-used kernel based change point detection method~\cite{truong2020selective}. 

To test the robustness of MtChD against noise, we generate synthetic data to mimic a scenario where there is one change point splitting the overall timeline into two, each of which consists of five days. We assume that the distribution of events remains the same within each interval. Three events before March 17 from the reference events are sampled for the first interval, and another three events that occurred after March 17 are sampled for the second interval. Additionally, we add noise events randomly selected from the noise pool at a specified ratio. We use the same noise pool for both intervals to ensure consistency. 

To test the robustness of MtChD against event overlaps, we adopt a similar process, but with increasing amount of overlap between periods. we randomly select one event from the reference that is not one of the six events selected before as the overlap event. We specify the overlap ratio $a_1$ for interval one, the overlap ratio $a_2$ for interval two, and the number of days where the overlap events occur. We randomly select narratives from the pool that represent the overlap event, and add them to the data.

\subsubsection{Evaluation on SRL clustering}

%Before jumping into the details of evaluation, we want to explain why we focus on the clustering performance on SRL triplets. One challenge of applying SRL triplets alone to present narratives is that a narrative could be captured in varied shapes, which makes the interpretation favors narratives that have only one form and therefore render the result to be less reliable. The traditional way of coping with the issue is to focus on limited semantic roles~\cite{jing2021characterizing}, yet it limits the comprehensive understanding of a large corpus. Therefore, one major contribution of our paper is to propose a way to aggregate SRL triplets with high semantic similarity. And given that SRL models have been wildly adopted (e.g., \cite{tangherlini2020automated,shahsavari2020conspiracy, zhong2019reasoning}) and evaluated (e.g., \cite{he2017deep,shi2019simple}), and we already tested the possibility of treating a triplet as a short sentence and feed it into SentBert, here we only focus on evaluating the performance of BIRCH in aggregating highly semantic similar SRL triplets in this section. 

We adopt the idea of \emph{coherence} from dynamic topic modeling to evaluate the quality of predicted clusters, which is the product of BIRCH on narrative fragments. 
In essence, the idea is about quantifying how coherent or homogeneous each topical cluster is. 
As a metric, it is common to use Umass~\cite{mimno2011optimizing}. 
However, Umass focuses on the extent of \emph{word} overlap within predicted clusters, which does not necessarily capture semantic similarity. 
Therefore we define an embedding-based coherence measure, which use cosine similarity as a proxy of semantic similarity. We train all narrative fragments with a pre-trained SentBert model (``all-mpnet-base-v2''). And with the embeddings, we assign the average value of cosine similarity of all pair-wise narrative fragments within a predicted cluster. The measurement is applied to both Twitter corpora.

\begin{figure}
\centering
\includegraphics[width=0.75\columnwidth]{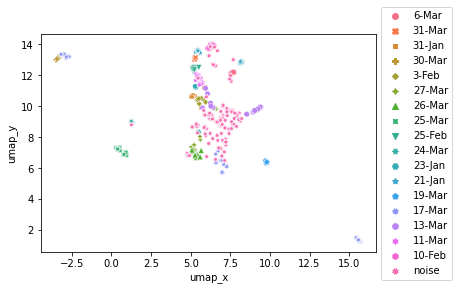}
\caption{UMAP of ground truth clusters and noise events}
\label{umap}
\end{figure}

We also evaluate the robustness of BIRCH against noise on synthetic data. All synthetic data generated for this task consist of 197 ground truth narratives from major events related to COVID, and certain amounts of randomly generated noise events (size ranges from 0 to 104). We also test the impact of BIRCH threshold on the performances. Specifically, we experiment with six different threshold values: 1, 1.5, 2, 2.5, 3, and 3.5. Fig~\ref{umap} shows the clear cluster structures of ground truth narrative, while noise events scatter in the space. 

Our evaluation uses two types of criteria: relative and absolute.
The \textbf{relative criterion} focuses on the pair-wise semantic similarity within a group. Narratives are considered accurately grouped if they belong to the same event and share a common cluster.
By contrast, the \textbf{absolute criterion} measures the ability of the clustering method to group similar narrative fragments together. We check whether the cluster with the most narratives belonging to a certain event matches the corresponding ground truth cluster.

\section{Results}
We demonstrate how our pipeline\footnote{check detailed tables and interactive figures at \url{https://osf.io/3aguz/?view_only=704d701a57224a3989e79b4b7726ceba}} works with two Twitter corpora (COVID-19 and French Election), and its robustness against noise and event overlaps with synthetic data. 

\begin{figure}[t]
\centering
\includegraphics[width=0.8\columnwidth]{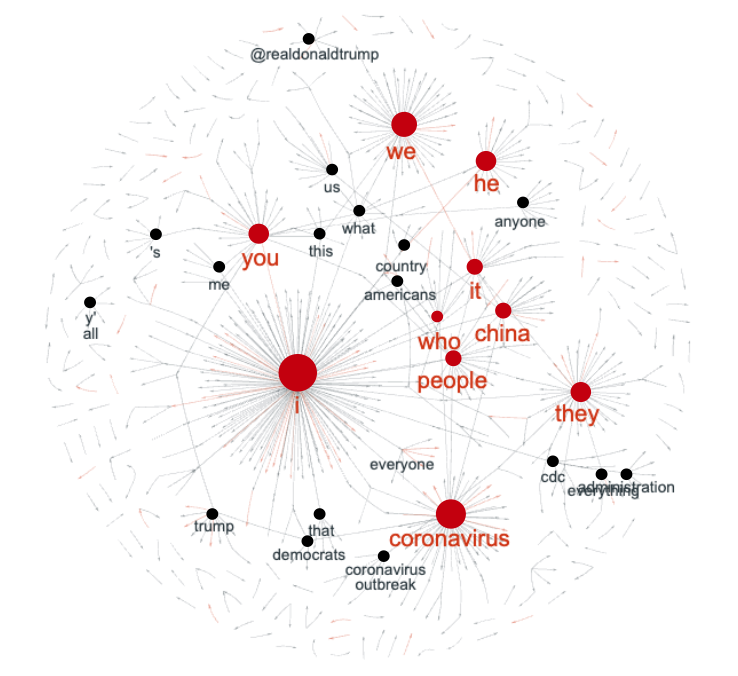}
\caption{The global narrative network of the COVID data.}
\label{Covid_global}
\end{figure}

\begin{figure*}[th!]
\centering
\includegraphics[width=1\textwidth]{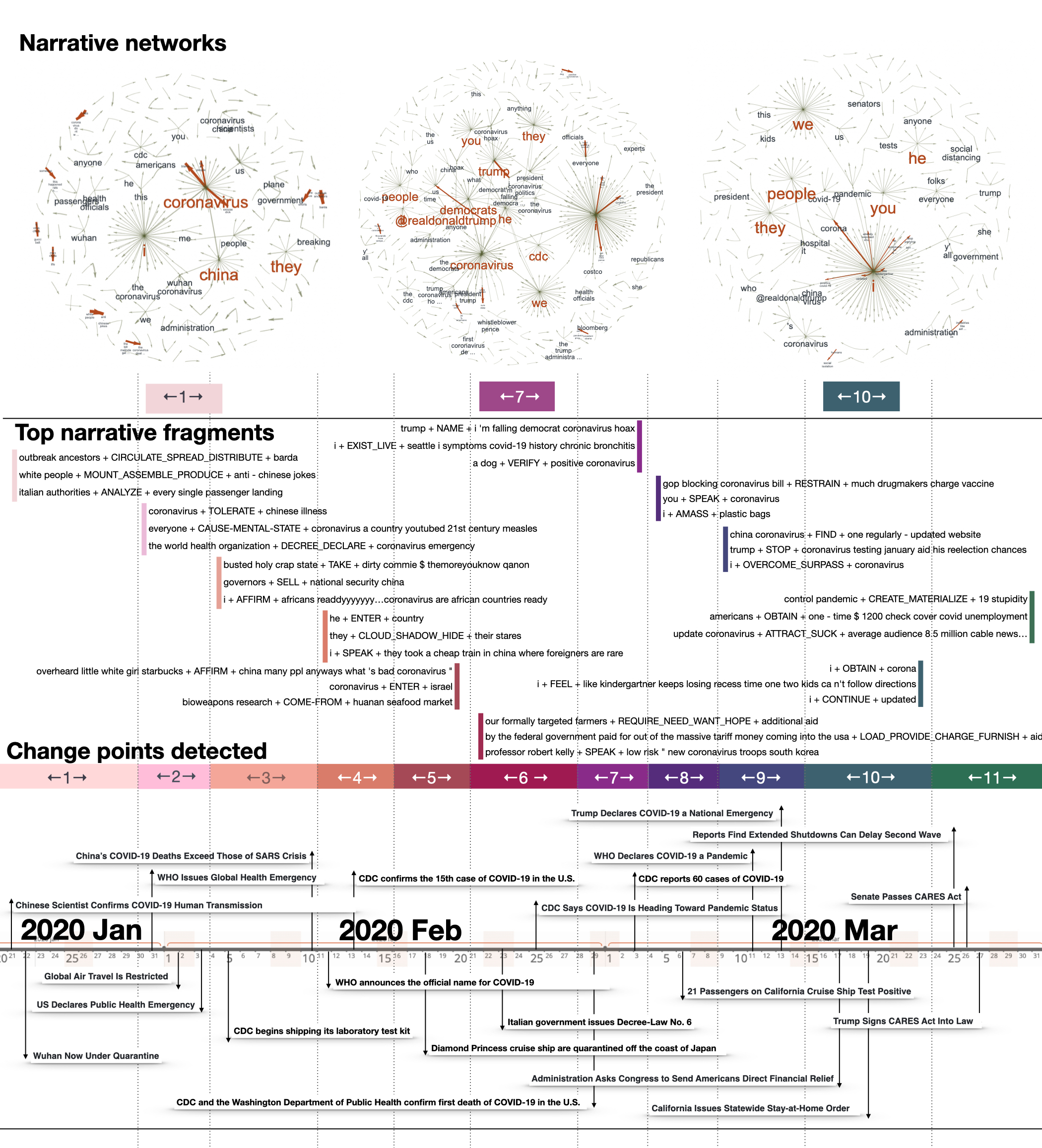} 
\caption{Local narrative networks extracted from the COVID-19 corpus. We set nodes with degree at least 10 in red, and node with degree between five and 10 in black. Orange edges represent the top narrative fragments in the backbone. The bottom subplot shows the timeline related to COVID provided by CDC and AJMC. The other subplots show results of change point detection, top three overall narrative fragments, and selected narrative networks respectively. \\ }
\label{Covid_local}
\end{figure*}

\begin{figure*}[th!]
\centering
\includegraphics[width=0.8\textwidth]{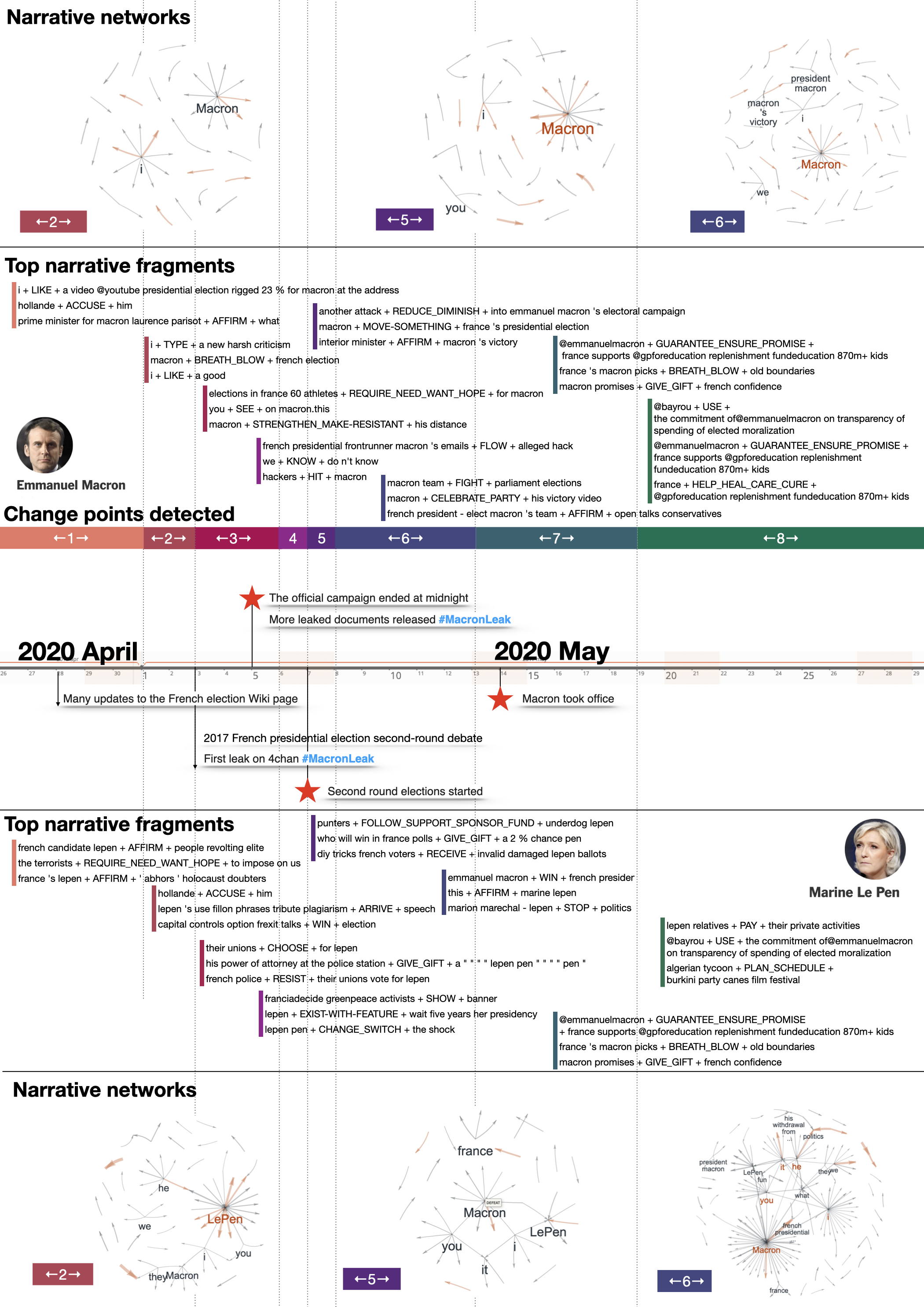} 
\caption{Local narrative networks with the 2017 French election data. Node and edge colors represents the same characteristics as in the previous figure. The bottom subplot shows the timeline from Wikepedia. The other subplots show results of change point detection, the top three narrative fragments, and selected narrative networks respectively.}
\label{French_local}
\end{figure*}

\subsection{COVID Data}
\subsubsection{Change points}
Figure.~\ref{Covid_local} shows the significant change points discovered, along with the timeline and narrative networks. The change point 02/28 is the most significant and it is close to the date of the very first US COVID-19 death. The next most significant change points are 02/11 and 03/04. The second change point coincides with California's declaring a state of emergency (03/04/20). Given that exploring details of the COVID data across all levels is beyond the scope of our study, here we only show results that are about micro-level narrative discovery, partitioning the data on significant change points: 01/30/20, 02/04/20, 02/11/20, 02/16/20, 02/21/20, 02/28/20, 03/04/20, 03/09/20, 03/15/20, and 03/24/20, resulting in 11 time frames. (see Figure.~\ref{Covid_local}). There are 35,650, 34,795, 25,968, 21,206, 21,086, 135,968, 42,872, 28,423, 27,819, 38,296, 39,059 unique narratives observed in the 11 time frames respectively. Please see the COVID event tree\footnote{\url{https://osf.io/3aguz/?view_only=704d701a57224a3989e79b4b7726ceba}} for more detail.

%For example, we could explore the macro-level narratives by comparing data from 01/21/20--02/27/20 with data from 02/28/20--03/31/20 (partitioning on the date of the first US COVID-19 death). Alternatively, we could explore the micro-level narrative shifts by checking continuous time frames shown in the lower level. 

\subsubsection{Top narrative fragments}

The top 15 representative narrative fragments are selected for the time frames (top three are listed in Figure.~\ref{Covid_local}). We find that ``anti-chinese jokes,'' ``China,'' ``Chinese,'' ``Wuhan,'' ``Wuhan virus,'' and ``Huanan seafood market'' were frequently mentioned from 01/21/2020 to 02/20/2020. The related narrative fragments change from ``coronavirus outbreak'' to ``Wuhan doctor + PERFORM + alarm coronavirus dies'', and later about ``bioweapons research from seafood market/super lab in Wuhan'' and ``concerns outbreak's economic impact.'' Besides China, other countries such as Israel and South Korea are also the top mentioned entities between 02/16/2020 to 02/27/2020. The United States is first frequently mentioned in 02/11/2020--02/15/2020 about ``horror coronavirus + REACH + san antonio'' and later about ``administration + AUTHORIZE\_ADMIT + infected americans'' and ``60 cases + CLOSE + us'' in 02/28/2020--03/03/2020. Interestingly, after 03/03/2020, narrative fragments related to countries lose their significance, and instead, networks are overtaken by narrative fragments starting with pronouns such as ``I,'' ``You,'' and ``He.''
This may be due to the surge of cases across the glob and signifies the pandemic phase. 

\subsubsection{Global narrative network}
We apply the backbone method on the global narrative network, extracting 924 nodes and 797 edges. We highlight the hubs (with degree at least 10) in red, and highlight edges which represent top 15 overall narratives (see previous section) in orange. As shown in Fig.~\ref{Covid_global}, more than half of the hubs are pronouns, including ``They'', ``He'', and  ``We''. Other hubs are more content-specific, including ``china'', ``people'', and ``coronavirus''. 
Among these hubs, most of the top narratives are related to ``People'', ``I'' and ``We''. In fact, there are 52 out of 92 top narrative fragments connected to hubs. The rest of the top narrative fragments tend to be isolated such that their end nodes have a degree of one. Examples include ``white people + MOUNT\_ASSEMBLE\_PRODUCE + anti - chinese jokes'', ``south korea + INFORM + 160 new cases'', and ``grocery stores + BEGIN + clearing out''.

\subsubsection{Local narrative networks}
According to Fig.~\ref{Covid_local}, starting from 01/21 to 02/27, ``China,'' ``Coronavirus'' and pronoun ``I'' are the main hubs in local narrative networks. Later, although ``Coronavirus'' and ``I'' remain as hubs, ``China'' gradually loses its significance. Starting from 03/04, pronouns such as ``We,'' ``They'' and ``He'' along with ``I'' start to appear as new hubs. After 03/15 they even have higher significance in the local networks compared to the ``Coronavirus.'' Furthermore, the overlap between local networks and the top 15 overall narratives suggest that on average, eight top narratives are captured by local networks. 

\subsection{French Election Data}

\subsubsection{Change points} 
The most significant change point in the French Election data corresponds to the second round of voting (May 7th, 2017), which pitted Emmanuel Macron against Marine Le Pen in the runoff elections. The next-level change points correspond to the start of the \#MacronLeaks campaign (05/03) and Macron's inauguration, which took place on 05/14. 

%We can split data into two time frames---data from 04/26 to 05/06 and from 05/07 to 05/28. Alternatively, we could dissect the data on the third level change points, including 05/03, 05/07, and 05/13. 

\begin{figure}
\centering
\includegraphics[width=0.6\columnwidth]{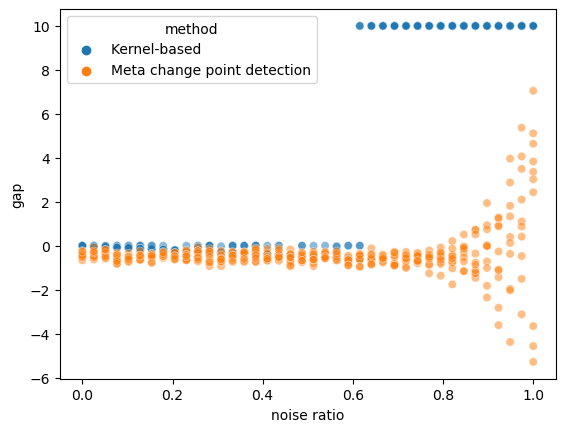}
\caption{Robustness of Meta change point detection against noise. The gap refers to the distance between the ground truth change point and the predicted one. The plot shows the predictions by MtChD are pretty stable until noise ratio reaches 0.8; while the performances of the Kernel-based method drastically drop when noise ratio is larger than 0.6.}
\label{cp_noise}
\end{figure}

Similarly, the results generated by MtChD includes change points with multiple granularity. 
The following results are based on the most fine-grained change points, which include 05/03, 05/05, 05/07, 05/08, 05/13, and 05/19, resulting in eight time frames (See Fig.~\ref{French_local}). There are 153,339, 68,297, 118,973, 65,281, 79,122, 126,446, 83,031, 63,617 unique narratives found in the eight time frames respectively.  

\subsubsection{Top overall narrative fragments}
From the top 15 narrative fragments related to Le Pen, we see those related to her political ideology. For example, ``people revolting elite'' is the most representative narrative fragment from 04/26 to 04/30, ``euro deadweight capital controls option'' is repeatedly shown as the top narrative fragments in 05/01--05/02, and in 05/03--05/04. During the same time, narrative fragments with emotions keep appearing, such as ``russia sanctions'' and ``holocaust denial'' found in 04/26--04/30, ``use fillon phrases tribute plagiarism'' in 05/01--05/02, and ``french police + RESIST + their unions vote for lepen'' in 05/03--05/04. Later staring from 05/05, rumors about ``invalid damaged lepen ballots'' start to spread. And the top narrative fragments after 05/08 shift the attention to the winning of Emmanuel Macron although the original tweets include keywords related to Le Pen. 

On the other hand, the top narrative fragments about Macron are mostly positive. For example, the endorsement from Laurence Parisot, Melenchon, and athletes were the main narrative fragments in 04/26--04/30, 05/01--05/02, and 05/03--05/04 respectively. In 05/05--05/06, seven narrative fragments concern with ``Macron Leak,'' an anti-Macron influence campaign~\cite{vilmer2021fighting}. Although they still remain heated in the next interval, narrative fragments shift to Macron winning the election. And starting from 05/08, the major narrative fragments are about ``macron team + Fight + parliament elections'' (No.1 in 05/08--05/12), Macron's political moves such as support "\@gpforeducation" (No.1 from 05/13--05/18), and ``transparency of spending of elected moralization'' (No.1 from 05/19--05/29). These narrative fragments reflect important priorities of Macron's administration, after he was sworn into office (05/14).

\subsubsection{Core narrative networks}
The local networks related to Le Pen and Macron exhibit big differences. In local networks related to Le Pen, node ``Macron'' gains more and more attention starting at 05/01, and after 05/07 it attracts more attention compared to node ``LePen'' and even takes over the core position starting 05/08. However, we don't see similar phenomenon in local networks related to Macron. Node ``Macron'' always maintains the core position and node ``LePen'' is never the hubs. In term of the overlap with top narratives, on average nine out the 15 top narratives are captured by local networks related to Le Pen, and on average 11 out of 15 top narratives are captured by local networks related to Macron. The comparison helps us understand the difference between statistical measurement and network-based method in uncovering significant narrative fragments.

% new
\subsection{Evaluation}

\begin{figure}
\centering
\begin{subfigure}[t]{1\linewidth}
    \centering\includegraphics[width=\columnwidth]{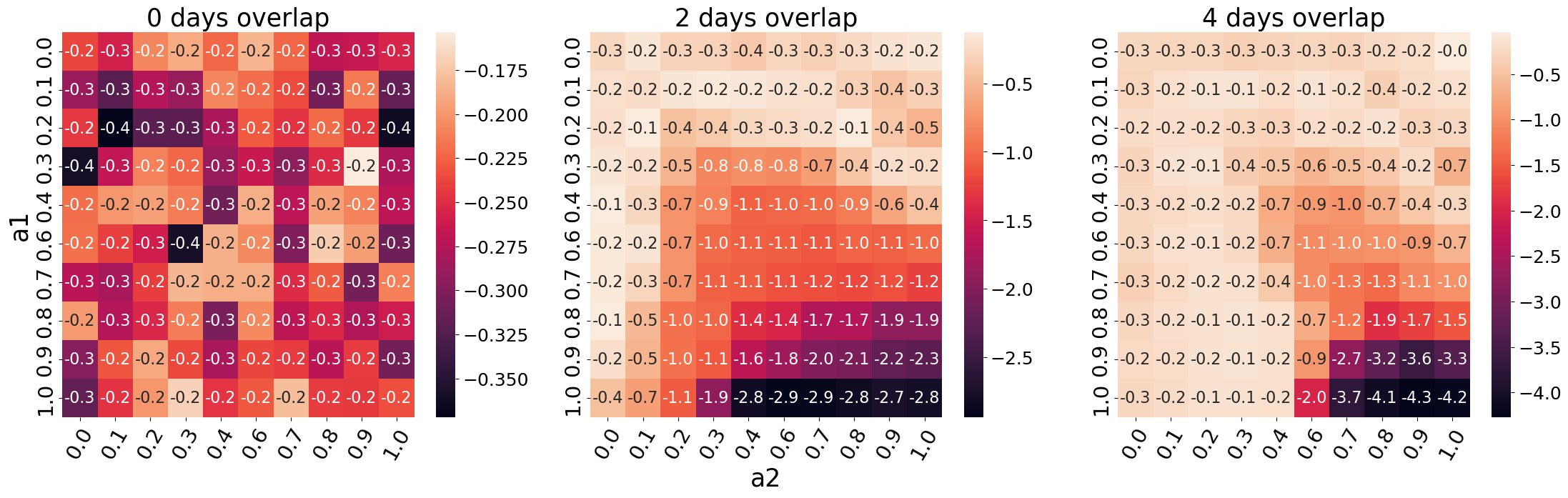}
    \caption{Meta change point detection}
  \end{subfigure}\\
\begin{subfigure}[t]{1\linewidth}
    \centering\includegraphics[width=\columnwidth]{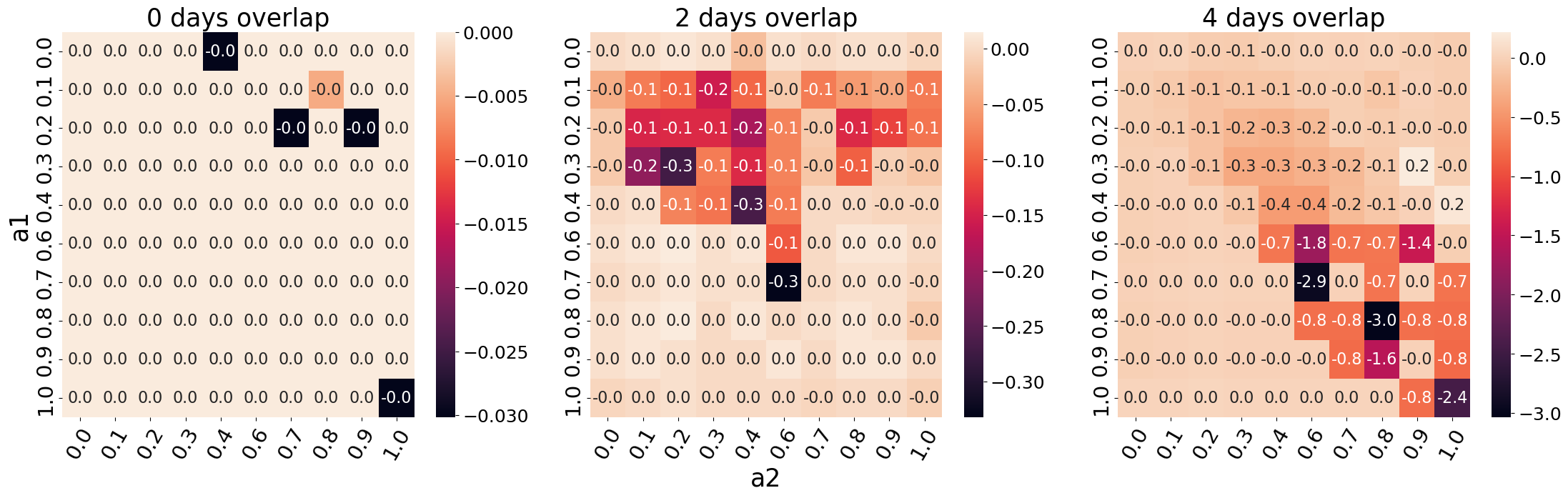}
    \caption{Kernel-based change point detection}
\end{subfigure}
\caption{Robustness of change point detection methods against event overlap. We generate synthetic data with one change point data. The x axis represents overlap ratio $a_1$ for interval one, and y axis is about overlap ratio $a_2$ for interval two. The value within a cell is the gap between ground truth change point and the predicted change point given a pair of $a_1$ and $a_2$. We show cases of 1) no overlap 2) 2-days overlap and, 3) 4-days overlap for both MtChD and the Kernel-based method. The performances of MtChD show association with $a_1$ and $a_2$. Similar phenomenon is not found in the performances by the Kernel-based method. }
\label{cp_overlap}
\end{figure}

\begin{figure}[t]
\centering
\begin{subfigure}[t]{1\linewidth}
    \centering\includegraphics[width=\columnwidth]{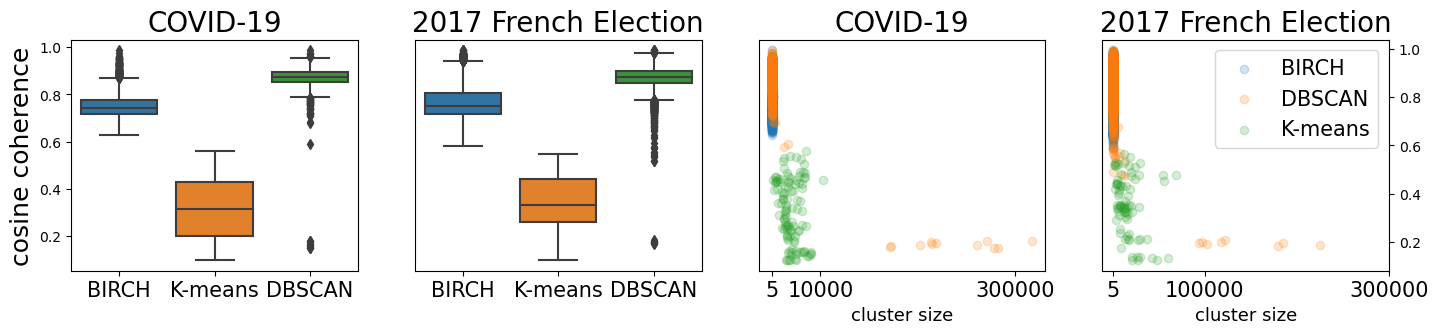}
    \caption{Empirical}
  \end{subfigure}\\
  \begin{subfigure}[t]{1\linewidth}
    \centering\includegraphics[width=0.8\columnwidth]{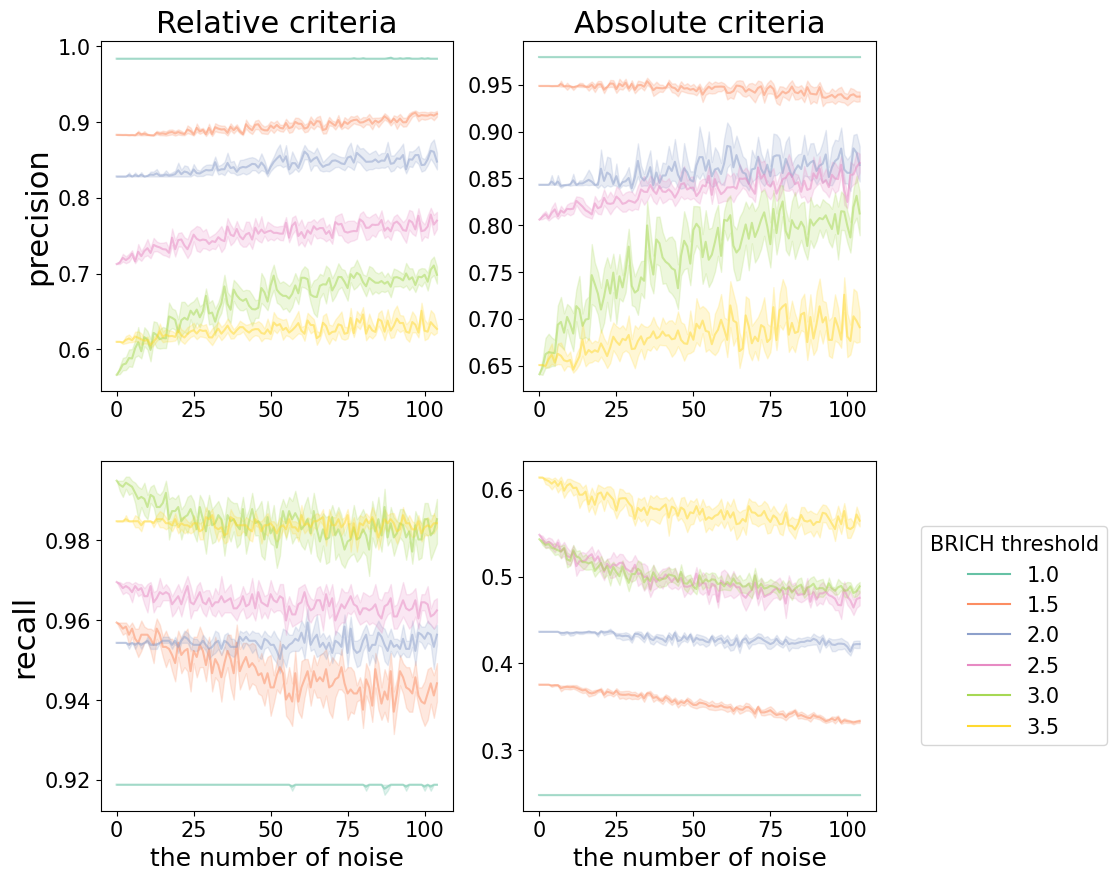}
    \caption{Synthetic}
  \end{subfigure}

\caption{Evaluation on SRL clustering quality. The two subplots on the left in (a) show the cosine coherence value indicating how BIRCH, DBSCAN, and K-means perform in clustering the two Twitter corpus--- the COVID-19 and the 2017 French Election data. The other two subplots on the right show the association between cosine coherence and the size of a cluster. Subplots in (b) show the precision and recall of BIRCH model in clustering synthetic data under relative and absolute criteria. In addiction, we show the changes in performance as the number of noise rises.  }
\label{srl_evaluation}
\end{figure}

\subsubsection{Robustness of Meta change point detection}
In both experiments, performance is evaluated by the distance between true change point, on day five, and the predicted one. Fig.~\ref{cp_noise} shows the gap between ground truth and the predicted change point. Meta change point detection (MtChD) remains accurate until noise ratio reaches 0.8, while the Kernel-based method~\cite{truong2020selective} shows worse performance  after noise ratio of 0.6, showing weaker robustness against random noise. 

Additionally, we find that accuracy drops as the number of overlap days increases in predictions by both methods (Fig.~\ref{cp_overlap}). Moreover, as $a_1$ and $a_2$ increase, we observes the gap between group truth and predicted value grows in the results of the Meta change point detection. And in extreme cases, where $a_1$ and $a_2$ equal to one, the gap indicates high reliability by predicting the date earlier than ground truth by the amount of overlap days. This shows that MtChD is good at capturing systematic changes in data. However, we do not see similar pattern in the results of the kernel-based method, where the gap does not seem to have associate with $a_1$ and $a_2$. In fact, the results of 2-days-overlap and 3-days-overlap do not make sense, by predicting day five as the candidate change point when $a_1$ and $a_2$ close or equal to one.  Once again, our method shows higher robustness to event overlaps compared to the Kernel-based method.

\subsubsection{Evaluation of narrative clusters}
With the evaluation on the Twitter data, we find that both DBSCAN and BIRCH could generate clusters with high average coherence (around 0.8) compared to K-means, but the results of DBSCAN have more outliers with very low coherence (Fig.~\ref{srl_evaluation} (a)). 
By the nature of the coherence measure, it may be reasonable to expect that the larger clusters would have lower coherence. 
This can have an outsized impact on evaluation results if one method produces larger clusters than the others. 
To analyze this possibility, we check the relationship between cluster size and coherence.
The result shows that all clusters generated by BIRCH have coherence value no less than 0.6, and their sizes are less than 10,000. Although most of the clusters predicted by DBSCAN have coherence value above 0.6 and decent size, yet a couple of the clusters are huge and exhibit low coherence. It suggests huge clusters with extremely low coherence generated by DBSCAN would introduce non-trivial bias for the downstream task---calculating the significance of the aggregated narrative fragments. In sum, these results suggest BIRCH produces more reliable clustering. 

In addition, our evaluations (Fig.~\ref{srl_evaluation} (b)) on synthetic data demonstrate that precision of BIRCH are all above or at least around 0.6 under both relative and absolute criteria, and the values further increase as more and more noise events are added. Recalls under relative criterion are all above 0.92, and they have slight decline as the number of noise rises. Although recalls under absolute criterion are not as high as that in relative criterion, they remain stable against the increase of noise. In sum, our method is robust against noise. 

\section{Discussion}
Here, we propose a framework to automatically extract narratives from a stream of tweets. By combining change point detection with SRL, we investigate the collective narrative shifts. We also evaluate and explore the performance of our framework. 

Core narratives captured by both methods align with the timeline provided by AJMC \footnote{\url{https://www.ajmc.com/view/a-timeline-of-covid19-developments-in-2020}} and CDC \footnote{\url{https://www.cdc.gov/museum/timeline/covid19.html}} about COVID, and that provided about French Election 2017\footnote{\url{https://en.wikipedia.org/wiki/2017_French_presidential_election}}, suggesting the substantial feasibility of our method.

\subsubsection{Temporality is essential}

Comparing the global narrative network with local narrative networks yield some notable insights. First, most of the hubs are highlighted in at least one of the local networks. However, without the clear time boundary pinpointed by change point detection, the persistence of the hubs is unclear. For example, the entity ``China'' is the core node before 03/03, but gradually loses attention, as the COVID-19 moves from the local epidemic in China to the pandemic. And although ``He,'' ``You,'' and ``They'' are shown as the hubs in the global network, they mainly have not become hubs until 03/04. Without change point detection, it would be challenging to uncover such narrative shifts. 

\subsubsection{Ambient storytelling} One hub that catches our attention concerns the entity ``Trump.'' Despite seldom appearing as a hub in local networks, it attracts plenty of attention between 02/28 and 03/14. This can be considered as evidence of \emph{ambient storytelling}~\cite{page2013small,sadler2021fragmented}---collective narrative fragments with no strong sequel or causal link yet building up a complete narrative. Taking a closer look at narrative fragments connect to ``Trump'' in the global network, we find ``trump + STOP + coronavirus testing January aid his reelection chances,'' ``trump + AFFIRM + coronavirus outbreak 'all control' 'very small problem','' ``trump + DISMISS\_FIRE-SMN + government's entire pandemic response chain command including white house management infrastructure,'' and ``democrats + ACCUSE + trump'' forming a narrative about Trump's role in controlling COVID-19 pandemic. 

In addition to narrative fragments related to the core entities, we also find another type of ambient storytelling, which has direct links between narrative fragments. For example, the combination of ``administration + FOLLOW\_SUPPORT\_SPONSOR\_FUND + people'' and ``people + REQUIRE\_NEED\_WANT\_HOPE + stay home''  in local network 03/15--03/23 is a clear narrative about officials ordering lockdowns in their jurisdictions. 

\subsubsection{Horizontal storytelling} Other than ambient storytelling, we also find evidence of \emph{horizontal storytelling}~\cite{sadler2021fragmented, bal2009narratology}, where one piece of narrative fragment could imply the whole story. For example, in the local network that depicts narratives related to Le Pen in 05/03--05/04, ``students + SPEAK + vote against lepen'' is a standalone narrative. Similarly, in the 05/05--05/06 local network about Macron, the narrative fragment ``hackers + HIT + macron'' is a complete summary of the \#MacronLeaks incident~\cite{vilmer2021fighting}. 

Among all horizontal storytelling, adding to the previous research on Trump's tweets~\cite{sadler2021fragmented}, we find evidence of narrative fragments using technique of \emph{sideshadowing}~\cite{morson1994narrative}, which attracts viewers' attention by emphasizing what \emph{could have} happened. For instance, ``diy tricks french voters + RECEIVE + invalid damaged lepen ballots'' suggests an anti-Macron conspiracy, misleading Le Pen's supporters to believe that election fraud prevented their candidate from winning. Likewise, ``russia sanctions + PRECLUDE\_FORBID\_EXPEL + lepen election win.'' is a sensational accusation without a proof, leaving room for imagination that Le Pen were suppressed by foreign behind-the-scene forces. 

\subsubsection{Narrative network or narrative fragments?}
One thing that captures our interest is a strong overlap between the top overall narratives and local networks. The average overlap rate is 10/15 for the COVID, and 9/15 and 11/15 for Le Pen- and Macron-subsets respectively. Among the overlapping narratives, few are connected to hubs. In some extreme cases, for example, the largest hub,  ``Macron'' in Le Pen local network 05/07, is not involved in any of the top 15 narratives in the same time frame. This suggests that hubs might have a larger probability of being involved in important narratives than non-hub nodes, yet, whether a narrative could be representative is not determined by its connection to a hub in a narrative network. 

At this point, however, we could not draw a definite conclusion that the network-based method outperforms the statistical method in narrative discovery due to the absence of the benchmark dataset about narratives. Nevertheless, we could use audiences' adoption of narratives as a proxy for evaluation purposes. A previous study has shown that the ``persistence'' of a tweet, which is quantified as the repeated exposure under a specific topic, has a significant marginal effect on audiences' adoption~\cite{romero2011differences}. Building on this idea, we could consider the frequency of a narrative fragment as the measure of its importance, and further refine it with measures like log-odd ratio with informative Dirichlet priors~\cite{monroe2008fightin}. This way, the top representative narrative fragments would be the potential reference data for narrative networks comparisons. 

If so, the network-based method would suffer from limitations of capturing essential narrative fragments, resulting in narrative mismatches. With our data, all mismatched top narratives connect to at least one long entity. For example, we find ``nigel farage formally endorses + EXIST--WITH--FEATURE + a lot say about lepen european union'' and ``we academics and researchers + SHOW + our support for emmanuel macron'' in the French Election data, and ``billionaire jack ma + CREATE\_MATERIALIZE + coronavirus vaccine roughly equivalent average u.s . family'' in the COVID data. Such long entities might have a low value of degree and are therefore hard to satisfy the filtering criteria based on backbone disparity.

Since the essential information of a narrative is stored in the edges in a narrative network, the edges carry information about how representative a narrative fragment is and what the relationship between the end nodes is. Therefore, by applying community detection, we overlook the exact meaning of the relationship attached to the edges and considering all edges equal. Especially in the case where competing narratives co-exist in a narrative network, such a move is not trivial. For example, when applying random walk (e.g., PageRank~\citealt{brin1998anatomy}) to capture importance of nodes, it is not reasonable to assign equal probabilities to edges ``I + AGREE + vaccine'' and ``My baby + NON-TOLERATE + vaccine'' simply based on the fact that they share both edge weights and next pathway to ``vaccine + KILL + children'', because individuals who endorse the narrative ``I + AGREE + vaccine''  would hardly consider ``vaccine + KILL + children'' acceptable, not to mention collectively forming a narrative linking the two fragments. Similarly, it is rarely likely that a collection of narratives with contrasting beliefs would be considered as two communities, instead of one, when applying community detection. That said, it should be noted that we do not intend to claim that applying node-based network methods is incorrect. Instead, we encourage the researchers to try these methods when contrasting narratives is not the main focus of the study (e.g.,~\citealt{shahsavari2020conspiracy}).

\section{Conclusion}
In summary, this study contributes the following. First, We propose an automated pipeline that identifies the evolution of collective narratives with statistical and network-based methods. The pipeline has successfully identified the major events and the collective opinions related to major entities. We have quantitatively evaluated the robustness of our method with synthetic data. While working with the pipeline, we find evidence in support of various forms of storytelling emphasized in traditional studies of narratives. 
We also discuss the difference between statistical measurements and network analysis. We acknowledge the strengths of network-based methods in detecting core elements of narratives, and their capability in retaining major representative narratives even after backbone filtering. Nevertheless, we caution the use of network-based methods when analyzing contrasting narratives. Future efforts could be put to test the general applicability of our approach in data with competing narratives and explore more sophisticated network-based methods to study narrative networks. 

%\subsubsection{Ethical statement.}
\subsubsection*{Broader perspective, ethics and competing interests}
We analyzed tweets discussing COVID-19 pandemic and French Elections, but we expect our narrative discovery methods to generalize to other datasets.
All data is public and collected following Twitter's terms of service, with the study considered exempt by the authors' IRB. To minimize risk to users, all identifiable information was removed and analysis was performed on aggregated data. Only public entities and politicians are mentioned explicitly by name in the data. Authors declare no competing interests.

% add it when paper is accepted
% \subsubsection*{Acknowledgement}

\bibliography{aaai22}
%\subsubsection{Acknowledgments.}

%\subsubsection{Appendices.}
%\thispagestyle{plain}
\end{document}